\documentclass[a4paper]{article}

\usepackage{INTERSPEECH2022}

\usepackage{color}
\usepackage{todonotes}
\usepackage{tipa}
\usepackage{multirow}
\usepackage{url} 

\usepackage{xspace}

\newcommand{\numlanguages}{12\xspace}
\newcommand{\datasetname}{AfricanVoices\xspace}
\newcommand{\numcreated}{3\xspace}

\usepackage{romannum}
\usepackage[T1]{fontenc}
\usepackage{makecell} 
\usepackage{array}
\usepackage{hyperref}
\newcolumntype{H}{>{\setbox0=\hbox\bgroup}c<{\egroup}@{}}

\title{Building African Voices}
\name{Perez Ogayo$^\dagger$, Graham Neubig$^\dagger {^\ddagger} $, Alan W Black$^\dagger$}

\address{
  $^\dagger$Language Technologies Institute, 
  Carnegie Mellon University\\
  $^\ddagger$Inspired Cognition\\
  Pittsburgh, PA, USA
 }

\email{ \{aogayo, gneubig, awb\}@cs.cmu.edu}

\begin{document}

\maketitle

\begin{abstract}

Modern speech synthesis techniques can produce natural-sounding speech given sufficient high-quality data and compute resources. However, such data is not readily available for many languages. This paper focuses on speech synthesis for low-resourced African languages, from corpus creation to sharing and deploying the Text-to-Speech (TTS) systems. 
We first create a set of general-purpose instructions on building speech synthesis systems with minimum technological resources and subject-matter expertise. Next, we create new datasets and curate datasets from "found" data (existing recordings) through a participatory approach while considering accessibility, quality, and breadth. 
We demonstrate that we can develop synthesizers that generate intelligible speech with 25 minutes of created speech, even when recorded in suboptimal environments. Finally, we release the speech data, code, and trained voices for \numlanguages African languages to support researchers and developers.%

\end{abstract}

\noindent\textbf{Index Terms}: Speech Synthesis, Text-to-Speech, African Languages, Language Resources

\section{Introduction}
%

Speech synthesis modelling techniques are advanced such that it is feasible to achieve almost natural-sounding output if sufficient data is available \cite{tacotron2, synthasr}. Specifically, current TTS techniques mostly require high-quality single-speaker recordings with text transcription for at least 2 hours of speech \cite{8683862}. For widely spoken languages with many textual and audio resources, this kind of high-quality speech synthesis data is both readily available and relatively easy to create if it does not already exist \cite{black2019wilderness, blasi2022systematic}. However, the audio and text transcripts necessary to produce a deployable TTS engine are not easily obtained for many other languages worldwide.

In this paper, we focus on African languages, which despite often having a relatively large number of speakers, tend to lack high-quality speech synthesis data.
This state of resource scarcity across the continent is not unique to speech synthesis but encompasses the entire field of language technologies  \cite{arya_mccarthy_new_2017,nekoto2020participatory}.
Arguably, this has stemmed from many languages being overlooked due to greater economic incentives for other languages and the lack of African researchers and technologists in industry and academia \cite{blasi2022systematic}.

We describe the development of an initiative, \datasetname, that attempts to change this state of affairs through a participatory methodology to create and curate single-speaker speech synthesis datasets for African languages.
The following principles guide us:

\noindent \textbf{Accessibility}: An important consideration when building TTS engines for low-resource languages is the ease of access for those who need it.
To this end, we make all developed data and trained speech synthesizers publicly accessible under user-friendly licenses.
We also focus on underlying technology that is easy to train and deploy in low-resource environments, such as low-powered Android smartphones.
To do so, we build on top of the long-standing FestVox project \cite{anumanchipalli2011festvox} to build deployable TTS engines.
Our generated TTS models use the CMU Flite \cite{black01b} framework, using the random forest based statistical synthesizer \cite{rfs15} for voices that are directly deployable on all Android phones through the open Google TTS API.
In addition, data created during this process is also suitable for a wide range of current and future corpus-based synthesis techniques; thus, better synthesizers can be built with this data in future iterations. may be created with this data in future iterations.

\noindent \textbf{Quality}: We curate high-quality data for a few languages to maintain maximum quality while utilizing cost-effective methods.
Importantly, to allow data creation to be a participatory process,  we also encourage community input by providing comprehensive guidance on topics such as data collection and licensing.

\noindent \textbf{Breadth}: Additionally, we include found data from the web in the \datasetname dataset to cover many languages.
We follow the CMU Wilderness project \cite{black2019cmu}, which bootstraps datasets from found data (any long-form audio with its transcript) using initial cross-lingual acoustic models to get the initial alignment and then in-language acoustic models to improve the dataset.

We open-source \datasetname which includes a speech synthesis corpora for 12 languages (including \numcreated that we record) and accompanying ready to use speech synthesizers on the \datasetname
website \footnote{ Accessible at \url{https://www.africanvoices.tech/}}. We release alignment  code, and number dictionaries \footnote{Accessible at \url{https://github.com/neulab/AfricanVoices}}.

\section{Focus Languages} 

\begin{figure}[t]
\centering
\includegraphics[width=40mm, scale=0.4]{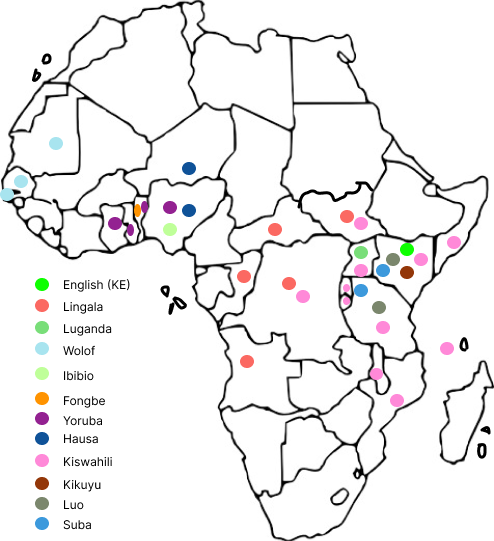}
\caption{Current language coverage of African Voices}
\label{fig:map}
\vspace{-5mm}
\end{figure}

\datasetname aims for both breadth and depth in its eventual goal of having high-quality voices for all African languages. The languages covered in our initial iteration of the dataset are spoken in Western, Central and Eastern Africa. Figure \ref{fig:map} shows the current language coverage.

\subsection{Created Data}
We curated text and recorded high-quality speech data for \numcreated languages: Luo, Suba and Kenyan English.

\textbf{Luo} (luo) (Kavirondo Luo/Dholuo) is a Nilo-Saharan language spoken by about 4.2 million speakers in Kenya and Tanzania.
It uses Latin script with 26 consonants and 9 vowels.
It is one of the most written languages in Kenya due to colonial and post-colonial politics and education policies~\cite{10.2307/45194720}.

\textbf{Suba} (sxb) is an endangered Bantu language spoken in Kenya and Tanzania. There are about 157,787 Kenyans who identify as ethnically Suba \cite{kenya_national_bureau_of_statistics_2019}; however, native speakers are less than 10,000 \cite{bernadine_racoma_olusuba_2014}. Most Suba speakers speak Dholuo as a
first language due to their assimilation into Luo culture, which
was accelerated by colonization \cite{10.2307/45194720}. 
Suba is written in Latin script. There are low literacy levels in Suba as Dholuo was the language of instruction in schools during colonial and early post-colonial periods. Consequently,
few written resources exist \cite{10.2307/45194720}. 

\textbf{Kenyan English} (en-ke) is the primary language of the government, media and schools and an official language alongside Kiswahili. English was introduced in Kenya through colonialism. Its vocabulary is similar to the British English, and local languages have heavily influenced its phonology \cite{nyaggah1990cross}.

\subsection{Found Data}
In addition, we used existing found data which were available under a permissive license. The languages below all use a modified version of the Latin script.

\noindent\textbf{Lingala} (lin) is a Bantu language spoken by about 45 million native and lingua franca speakers mainly in the Democratic Republic of the Congo and to a lesser degree in Angola, the Central African Republic and southern South Sudan \cite{8672222}.

\noindent\textbf{Kikuyu} (kik) is a highly agglutinative Bantu language spoken in the central region of Kenya by about 8  million people as a first language \cite{kenya_national_bureau_of_statistics_2019}. 

\noindent\textbf{Yorùbá} (yor) is a Niger-Congo language spoken by about 34 million speakers in Nigeria and other countries on the West African coast \cite{eberhard_ethnologue:_2022}. 

\noindent \textbf{Hausa} (hau) is an Afro-Asiatic language of the West Chadic branch with 60 million native speakers \cite{ethno2017}. It is spoken in Nigeria and Niger. 

\noindent \textbf{Ibibio} (ibb) is a Benue–Congo language with about 10 million speakers in Nigeria.

\noindent \textbf{Wolof} (wol) is a Niger-Congo language spoken by about 10 million speakers in Senegal, Gambia and Mauritania \cite{8672222} . 

\noindent \textbf{Luganda} (lug) is a Bantu language spoken in Uganda by about 20 million speakers.

\noindent \textbf{Fongbe} (fon) is a Niger-Congo language spoken in Benin by about 4.1 million speakers \cite{noauthor_building_nodate}. 


\noindent \textbf{Kiswahili} (swa) is a Niger-Congo language with about 200 million speakers \cite{the_conversation_2022}. It is a lingua franca in the Africa Great Lakes region and an official language for the East African Community states.

\section{Dataset Creation}

One philosophy of \datasetname is that it is participatory, enabling interested parties to easily create voices for the languages they speak or are interested in.
To this end, we have created a comprehensive set of directions (available at \url{https://github.com/neulab/newlang-tech/tree/main/speech-synthesis}) to allow even those without extensive experience to create data and voices and contribute them back to the dataset if they wish.
We briefly outline the processes below, illustrating tricks, difficulties, or pitfalls we encountered with our focus languages.

\subsection{Creating Speech Data}
\noindent\textbf{Developing Prompt Sets}
We collect textual data in the target languages by scraping various websites. Most of the data was created and published originally in our target languages. 
 The text data sources for Luo and Suba included website copy, folklore and stories, research papers and theses, grammar and instructional books, and social media text. Most of the Suba text was obtained from \cite{subalanguage}, a website part of a project to revitalize the language. Luo data included translated data from an English news corpus developed from news articles by Kenyan media houses. It was easier to collect data for Luo, where we obtained 13,879 than for Suba, where only 2,078 utterances were collected.

The lack of a standardized orthography for both languages posed challenges, as there were several instances of alternate spellings. Some Suba utterances also contained a mix of dialects. For the Luo prompt set, we used Festvox tools \cite{anumanchipalli2011festvox} to select 1500 utterances that represent the phonetic and prosodic contexts of the language from the entire text corpus.
This selection was not possible for Suba as its textual dataset was small, and thus we used the entire text corpus. English prompt was obtained from CMU ARCTIC databases which consist of 1150  phonetically balanced utterances \cite{kominek2004cmu}.  

\noindent\textbf{Speaker selection }When creating a corpus for TTS, a speaker (voice talent) should be fluent, literate, trained and familiar with voice recording \cite{47393}. Speaker selection was crucial for Suba as few fluent and literate speakers exist. We advertised the position on a social media group whose objective was learning the Suba language. We received interest from participants who sent sample writings and recordings but proceeded with the person recommended by an authority in the language. The voice talents were paid in cash and kind.

\noindent\textbf{Speech Recording }
In many cases, obtaining studio-quality data for resource-constrained languages is impossible. Luo and an initial version of English were recorded in a residential house using a smartphone and in-ear microphone. We re-recorded English in a studio to improve quality. We recorded Suba at a  local radio station for Suba. While we could have recorded in any other studio, it was essential to relocate the voice talent to Mfang'ano island to use the local radio station, which enabled us to receive feedback from native speakers on the quality of recordings and text.  We spent  2 hours,  4-hour sessions over 8 days, and a one-night session to record  English, Luo and Suba, respectively.

\noindent \textbf{Quality Control }
Before recording, we removed utterances with mixed dialects or changed to the selected dialect. After recording, we modified the prompt-set to reflect the actual content of the speech.
We power-normalized the recordings to minimize the variation resulting from recording in different sessions and prosodic inconsistencies to ensure consistent volume. 

Since most African languages exist in multi-lingual environments, everyday speech contains many borrowed words and constant code-switching. We faced the challenge of using a word's foreign or adapted pronunciation. For example, Luo speakers would pronounce the  English word 'fish' as  /\textipa{fis}/ instead of /\textipa{fIS}/ because of the absence of /\textipa{S}/  in Dholuo phonology. In this case, we let the speaker use the pronunciation that was most natural to them.

\subsection{Aligning Found Data}
For many low-resourced languages, the Bible is a major source of text and audio. In this project, we used the New Testament section of the Bible from  Bible.is \footnote{https://www.faithcomesbyhearing.com/audio-bible-resources/bible-is} for Suba and  Open.bible\footnote{https://open.bible/resources/} for the rest. 
\subsubsection{Text preparation}
To preprocess the text, we added chapter introductions and subtitles present in the audio but missing in the script. We also normalized numbers. To this end, we release \textbf{\datasetname number dictionaries} \footnote{Available at \url{https://github.com/neulab/AfricanVoices/tree/main/number_dictionaries}}  for all languages we focus on that can be used to normalize numbers.
\subsubsection{Speech preparation}
Audio obtained from the Bible sources was saved as chapters in mp3 format. To segment and align it to utterance level, we followed the CMU Wilderness project's \cite{black2019cmu} segmentation and alignment process. Alignment for New Testament data, which is $\approx$ 20 hours of speech, took a maximum of 5 days per language on a 16-CPU machine. Table \ref{tab_found_data} shows the resulting utterances.

\begin{table}[h!]
\caption{
\label{tab_found_data} Data from found sources.  \\ \scriptsize **Data is not released \vspace{-2mm}
}
\begin{tabular}{lcccH@{\hspace*{-\tabcolsep}}}
\hline
\textit{Language} & \textit{source} & \textit{No. utterances} & \textit{hrs} & \textit{MCD}  \\
\hline
Luo & Open.Bible & 11263 & 15.92 & 5.20 \\
Lingala & Open.Bible &  12957 & 27.52 & 4.33  \\
Kikuyu& Open.Bible &   10877& 17.72 & 5.19 \\
Yoruba & Open.Bible &  10978 & 18.04  & 4.61 \\
Hausa-M & CommonVoice &  518 & 0.62  &  6.95\\
Hausa-F & CommonVoice & 1938  &  2.3 &  6.29 \\
Luganda & CommonVoice& 2942 & 4.52 & - \\
Ibibio &  LLSTI  & 125   &   0.32 & - \\
Kiswahili & LLSTI & 426  & 0.53 &  - \\
Wolof & ALFFA  & 1000  & 1.2 &  -\\
Fongbe & ALFFA  & 542  & 0.33 & - \\
Suba** & Bible.is & 11971 & 24.82 & 5.40  \\
\hline
\end{tabular}
\vspace{-4mm}

\end{table}

\subsection{Found Data Sources}
In addition to Open.Bible and Bible.is data that we aligned, we obtained data from the following sources in utterance format:

\noindent \textbf{LLSTI}: The Local Language Speech Technology Initiative project developed TTS datasets for localization of speech technology. We obtained Ibibio~\cite{llsti-ibibio} and Kiswahili~\cite{llsti-kiswahili} by converting the publicly distributed lpc and res files to wav using Festvox tools.

\noindent\textbf{Mozilla CommonVoice} : We selected data from a single speaker with the most utterances for Luganda and Hausa.

\noindent \textbf{ALFFA}: ALFFA project~\cite{gauthier-etal-2016-collecting}  developed TTS and ASR technologies and data for Kiswahili, Fongbe, Wolof and Amharic. We selected a single speaker subset of the data for each language.

\subsection{Corpus License}
We limited our collection to distributable sources to allow for free redistribution of the \datasetname data and works. 
The data from open.bible is distributed under CC by SA license, allowing for sharing and adaptation as long as they are attributed and distributed under the same license. We cannot redistribute Suba data from Bible.is and we exclude it from the public repo despite using it in experiments. We release all works from AfricanVoices under CC by SA license.

\section{Experiments} 
\label{exeriments}
To evaluate the effectiveness of the creation and curation processes  mentioned above, we perform experiments with Luo and Suba, seeking to answer the following questions:
\begin{itemize}
    \item RQ1: Are the datasets sufficient to build a high quality synthesizer in the targeted languages?
    \item RQ2: How much curated data is necessary?
    \item RQ3: How does curated data compare with found data?
\end{itemize}

To answer the questions above, for both created and found data, we divided each into splits of 25 min, 50 min and 101 min (the largest amount of created data was 102 minutes for both languages. 
As mentioned previously, to allow those with less experience in speech technology to expand \datasetname to other languages, we built TTS systems using Festvox tools \cite{anumanchipalli2011festvox} due to their relative accessibility as they do not require expensive compute resources.

\subsection{Results and Analysis} 
For objective evaluation, we used the mean Mel Cepstral  Distortion (MCD) score \cite{kominek2008synthesizer}.%
\footnote{Previous work reports that an improvement in MCD of 0.12 is significant and recognisable to listeners \cite{kominek2008synthesizer}.}
Table  \ref{tab_MCD_splits} shows the results of the automatic evaluation.

\begin{table}[ht!]
\centering
\caption{\centering
\label{tab_MCD_splits} Objective evaluation using MCD (lower is better).\vspace{-2mm}
}
\begin{tabular}{clrrr}
\multicolumn{1}{l}{\textbf{Lang}} & \textbf{Source} & \textbf{25} & \textbf{50} & \textbf{101} \\ \Xhline{1.1pt}
\multirow{2}{*}{Luo}              & Found           & 4.73        & 4.73        & 4.65         \\
                                  & Created         & 6.49        & 6.45        & 6.37         \\ \hline
\multirow{2}{*}{Suba}             & Found           & 4.67        & 4.40        & 4.37         \\
                                  & Created         & 5.15        & 4.58        & 4.80 \\ \hline        
\end{tabular}
\end{table}

We conducted human evaluation, specifically preference and transcription tests, to obtain subjective scores. We advertised the call for evaluators on social media platforms such as WhatsApp and Slack workspaces. To conduct the listening tests, we used TestVox\footnote{https://bitbucket.org/happyalu/testvox/wiki/Home
}, an open-source web-based framework for running subjective listening tests~\cite{inproceedings}. 

\noindent\textbf{Preference test} We did A/B test to compare the synthesizer created using found and created data. In this task, the evaluators were asked to respond to \emph{Listen to the two audio clips below and select the one you prefer.} by selecting either \emph{A, B, or No difference}. Tables  \ref{tab_luo_ab} and \ref{tab_suba_ab} show the results for the A/B test.

\noindent\textbf{Transcription test} We also conducted transcription tests where the evaluators were asked to transcribe the audio. The lack of a standardized orthography for both languages was a major challenge for this task. 
The most common  `errors' made by evaluators were (i) whether to join an agglutinated word or not e.g. \emph{kawuononi vs kawuono ni} and (ii) similar sounds, especially the semi-vowel \emph{w}  and vowel \emph{u} e.g. \emph{dwe vs. due} and  (iii) whether to use a double vowel or not e.g. \emph{Mbeeri vs Mberi}. We found no significant difference in the results from the different data splits. Luo and Suba had an average  CER  of (5.85 found and 5.80 created) and (10.80 found and 13.78 created), respectively.

Our results answer the questions in section \ref{exeriments} as follows:
\begin{itemize}
   \item RQ1:  Both objective and subjective tests show that created and found data are sufficient to build a synthesizer.
    \item RQ2: We found that at least 25 minutes of curated data is needed. Recording less than 25 minutes might not be worth the effort and cost of preparing to record.
    \item RQ3:  The A/B tests show that curated data is comparable to found data despite their recording conditions. The evaluators consistently preferred output from created Suba.
\end{itemize}

\begin{table}[h!]
\centering
\caption{\centering
\label{tab_luo_ab} Preference test results for Luo \vspace{-2mm}
 }
\resizebox{\columnwidth}{!}{
\begin{tabular}{llrrrr}
\textbf{Split}            & \textbf{Evaluator} & \multicolumn{1}{l}{\textbf{Found}} & \multicolumn{1}{c}{\textbf{Created}} & \multicolumn{1}{c}{\textbf{Same}} & \multicolumn{1}{c}{\textbf{Best}} \\
\Xhline{1.1pt}
\multirow{2}{*}{25 min}   & Evaluator1         & 10                                 & 10                                   & 0                                          & \multirow{2}{*}{Found}            \\ 
                          & Evaluator2         & 11                                 & 8                                    & 1                                          &                                   \\ \hline
\multirow{2}{*}{50 min}   & Evaluator1         & 11                                 & 9                                    & 0                                          & \multirow{2}{*}{Found}            \\
                          & Evaluator2         & 13                                 & 5                                    & 2                                          &                                   \\\hline
\multirow{2}{*}{101  min} & Evaluator1         & 7                                  & 13                                   & 0                                          & \multirow{2}{*}{Created}          \\
                          & Evaluator2         & 6                                  & 11                                   & 3                                          &    \\ \hline                             
\end{tabular}
}
\vspace{-4mm}

\end{table}

\begin{table}[h!]
\caption{\centering
\label{tab_suba_ab} Preference results for Suba \vspace{-2mm}
}
\begin{tabular}{llrrr}
\textbf{Split}            & \textbf{Evaluator} & \multicolumn{1}{l}{\textbf{Found}} & \multicolumn{1}{l}{\textbf{Created}} & \multicolumn{1}{l}{\textbf{Best}} \\
\Xhline{1.1pt}
\multirow{2}{*}{25 min}   & Evaluator1         & 1                                  & 19                                   & \multirow{2}{*}{Created}          \\
                          & Evaluator2         & 10                                 & 10                                   &                                   \\ \hline
\multirow{2}{*}{50 min}   & Evaluator1         & 0                                  & 20                                   & \multirow{2}{*}{Created}          \\
                          & Evaluator2         & 9                                  & 11                                   &                                   \\ \hline
\multirow{2}{*}{101  min} & Evaluator1         & 3                                  & 17                                   & \multirow{2}{*}{Created}          \\
                          & Evaluator2         & 0                                  & 20                                   &          \\ \hline                        
\end{tabular}
\vspace{-3mm}

\end{table}

 While the subjective results are not significantly different, the objective scores are, especially for Luo. This difference might be explained by the sensitivity of MCD \cite{kominek2008synthesizer}, which measures the difference between the original waveforms and the synthesized ones. We recorded the Luo data with headphones in a residential setting, so the sound quality generated has a lot of variations and background noise. 

It is important to note that not all languages, voices, and prerecordings are equal. Some found data may be really good (a consistent speaker), and some found may not (e.g. the Bible.is data has quiet background music). Some speakers from created data may be better than others, so experiments may have to be done for the target language.  

\section{ Related Work}
Creating a high-quality speech synthesizer demands high-quality single-speaker
corpus \cite{DBLP:journals/corr/abs-2106-08468} unlike automatic speech recognition (ASR), which requires a diverse multi-speaker corpus to capture different accents, speaker
characteristics, and acoustic environments. The voice talents who record the speech are
usually highly trained, fluent, and have experience recording speech. In low-resource settings, finding such speakers is hard due to the low economic development levels. Furthermore, the cost required to record in studios and extensively collect textual data poses a significant challenge to building high-quality TTS corpora for low-resource languages. This necessitates innovative approaches for speaker selection, speech recording, and post-
processing of recorded audio. 

When creating a multi-speaker speech corpora for 11 South African languages, Niekerk et al. \cite{47393} recorded audio in low-cost environments like university buildings using laptops and cheaper microphones and applying audio processing techniques tools to control things like background noise. Common Voice \cite{ardila2020common} is a platform to crowd-source transcribed speech corpora, including African languages. While the platform and resultant corpora are helpful for speech technology research and development, most of the data is less suited for speech synthesis as it is multi-speaker and recorded in varied environments.

Even when  African speech is available, it is often not well transcribed. For low-resource languages, it is often the case that
there are (i) no transcriptions (audios/videos available online,
e.g. from media stations or vloggers), (ii) quality is not sufficient for TTS, (iii) not enough speech from one speaker or (iv)
the licence under which the corpus is released is limiting. In the
case of (i), manual segmentation and transcription require much human capital \cite{dargis-etal-2020-development}.

 Most high-quality speech corpora and consequently speech synthesizers for African languages are commercial. Most of the available ones represent a small fraction of the 2000 languages, with South African languages dominating because of government support through SADiLaR \footnote{https://sadilar.org/index.php/en/resources}. We found the following speech synthesis resources for African languages:

    \noindent\textbf{Existing high-quality speech synthesis corpora} Some of the high quality include: NCHLT speech corpus \cite{DBLP:conf/sltu/BarnardDHWB14}, Lwazi II corpus, Gamayun Coastal Swahili speech corpus\footnote{https://gamayun.translatorswb.org/data/}
     
     \noindent\textbf{``Found data''} Data in this category include those available online as part of other projects, e.g. audiobooks, entertainment and news, and data created/availed for ASR. Mozilla Common Voice, Gamayun Congolese Swahili~\cite{9342939},  Open.Bible, Bible.is,  A Kiswahili Dataset for Development of Text-To-Speech System~\cite{rono2021a} 

    \noindent\textbf{Publicly available speech synthesizers} These are TTS systems that are available for free. An example is  LLSTI~\cite{noauthor_local_nodate} 
    
    \noindent\textbf{Commercial synthesizers} These include Microsoft Text-to-Speech, Google API, Ajala AI, Inclusive solutions.


\section{Conclusion}
This paper describes creating a speech corpus from found and created data sources for low-resource languages with a limited budget. We outline the challenges of creating a voice for an endangered language and suggest ways to overcome them. We find that $\approx$1 hour of speech is sufficient for creating an average synthesizer, even when recorded in suboptimal conditions. We build TTS for \numlanguages languages and open-source models and speech corpora; available at \url{https://www.africanvoices.tech/}.

\datasetname is an ongoing project, and we invite contributors to cover more African languages. Future work should increase the geographic diversity of the corpus, and it will be desirable to focus on languages that are at the most risk of extinction. In addition, further work can be done to develop web and mobile applications that can be used to record voices by untrained voice talent, thus making the process more accessible.

\section{Acknowledgements}

This work was supported in part by the National Science Foundation under grant \#2040926 a grant from the Singapore Defence Science and Technology Agency.

\clearpage
\bibliographystyle{IEEEtran}

\bibliography{mybib}

\end{document}